\documentclass[sigconf,screen]{acmart}

\pdfoutput=1

\copyrightyear{2022}
\acmYear{2022}
\setcopyright{acmcopyright}\acmConference[MM '22]{Proceedings of the 30th ACM
International Conference on Multimedia}{October 10--14, 2022}{Lisboa, Portugal}
\acmBooktitle{Proceedings of the 30th ACM International Conference on Multimedia
(MM '22), October 10--14, 2022, Lisboa, Portugal}
\acmPrice{15.00}
\acmDOI{10.1145/3503161.3548346}
\acmISBN{978-1-4503-9203-7/22/10}

\usepackage{algorithm}
\usepackage{algorithmic}
\usepackage{xspace}
\usepackage{multirow}

\begin{document}

\title{CLOP: Video-and-Language Pre-Training with Knowledge Regularizations}

\author{Guohao Li}
\email{liguohao@baidu.com}
\affiliation{
  \institution{Baidu Inc.}
  \city{Beijing}
  \country{China}
}

\author{Hu Yang}
\email{yanghu03@baidu.com}
\affiliation{
  \institution{Baidu Inc.}
  \city{Beijing}
  \country{China}
}

\author{Feng He}
\email{hefeng07@baidu.com}
\affiliation{
  \institution{Baidu Inc.}
  \city{Beijing}
  \country{China}
}

\author{Zhifan Feng}
\email{fengzhifan@baidu.com}
\affiliation{
  \institution{Baidu Inc.}
  \city{Beijing}
  \country{China}
}
\author{Yajuan Lyu}
\email{lvyajuan@baidu.com}
\affiliation{
  \institution{Baidu Inc.}
  \city{Beijing}
  \country{China}
}
\author{Hua Wu}
\email{wu_hua@baidu.com}
\affiliation{
  \institution{Baidu Inc.}
  \city{Beijing}
  \country{China}
}

\author{Haifeng Wang}
\email{wanghaifeng@baidu.com}
\affiliation{
  \institution{Baidu Inc.}
  \city{Beijing}
  \country{China}
}

\renewcommand{\shortauthors}{Guohao Li et al.}
\newcommand{\spara}[1]{\smallskip\noindent\textbf{#1}.}
\newcommand{\etal}{\text{et~al.}\xspace}
\newcommand{\red}[1]{\textcolor{red}{#1}}

\begin{abstract}
  Video-and-language pre-training has shown promising results for learning generalizable representations. Most existing approaches usually model video and text in an implicit manner, without considering explicit structural representations of the multi-modal content. We denote such form of representations as ``structural knowledge'', which express rich semantics of multiple granularities. There are related works that propose object-aware approaches to inject similar knowledge as inputs. However, the existing methods usually fail to effectively utilize such knowledge as ``regularizations'' to shape a superior cross-modal representation space. To this end, we propose a {\bf C}ross-moda{\bf L} kn{\bf O}wledge-enhanced {\bf P}re-training ({\bf CLOP}) method with Knowledge Regularizations. There are two key designs of ours: 1) a simple yet effective Structural Knowledge Prediction (SKP) task to pull together the latent representations of similar videos; and 2) a novel Knowledge-guided sampling approach for Contrastive Learning (KCL) to push apart cross-modal hard negative samples. We evaluate our method on four text-video retrieval tasks and one multi-choice QA task. The experiments show clear improvements, outperforming prior works by a substantial margin. Besides, we provide ablations and insights of how our methods affect the latent representation space, demonstrating the value of incorporating knowledge regularizations into video-and-language pre-training.
\end{abstract}



\maketitle

\section{Introduction}
\label{sec:intro}

\begin{figure}[t]
  \centering
  \includegraphics[width=.48\textwidth]{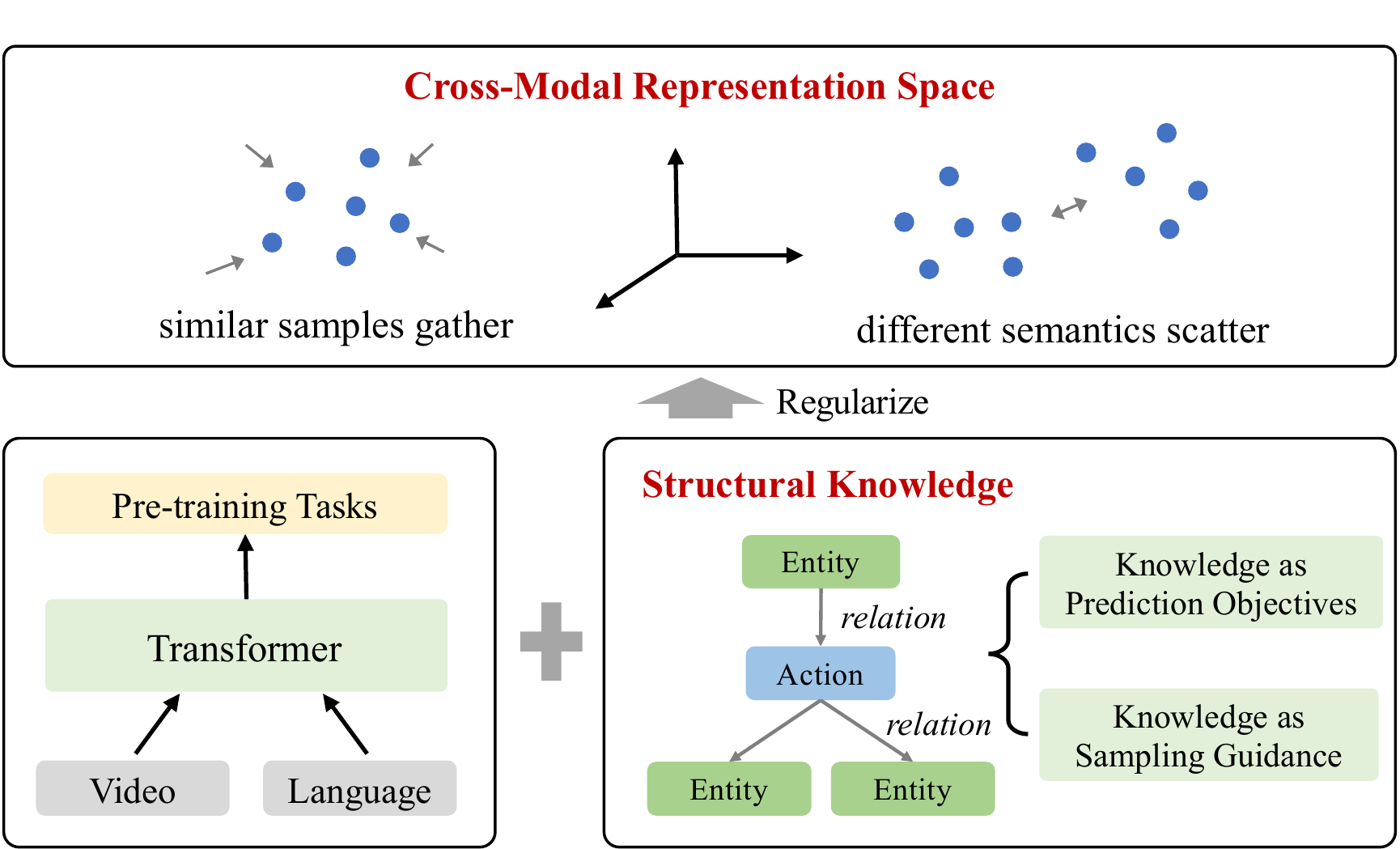}
  \caption{We incorporate structural knowledge to regularize the cross-modal representation space.}
  \label{fig:intro}
\end{figure}

Learning generalizable visual-and-language representations requires a comprehensive understanding of visual and textual inputs, as well as the semantic correlations between these modalities.
Recently, the research community has witnessed a burst of progress on image-text pre-training~\cite{lu2019vilbert,tan2019lxmert,li2019visualbert,li2020unicoder,chen2019uniter,yu2020ernie,li2020unimo,radford2021learning}, but pre-training for video-text is still in its infancy.

The dominant paradigm of video-text pre-training is to train powerful encoders (for video and text) with several self-supervised proxy tasks~\cite{sun2019videobert,luo2020univl,li2020hero,bain2021frozen}.
The canonical proxy tasks include ``conditioned masked modeling'' (predicting masked inputs from multi-modal context), ``video-text contrastive learning'' (discriminating positive samples from negatives), etc.
Essentially, the ultimate goal of this paradigm is to learn a scalable cross-modal representation space, where the video-text representations should reside properly and can be easily generalized to downstream tasks.

Most existing approaches usually model video and text in an implicit manner, without considering explicit structural representations of the multi-modal content.
We denote such form of representations as ``structural knowledge'', which express rich semantics of multiple granularities.
There are related works that incorporate such knowledge as inputs or masking constraints.
For example, the object-level regional features~\cite{anderson2018bottom} are widely used in image-text pre-training~\cite{lu2019vilbert,chen2019uniter,li2019visualbert,li2020oscar,yu2020ernie}.
The Ernie-Vil~\cite{yu2020ernie} incorporates knowledge by masking the semantical units in the text.
Several video-text pre-training works propose object-aware approaches to model visual and textual inputs~\cite{zhu2020actbert,wang2021object}.
However, the methods usually fail to utilize such knowledge as ``regularizations'' to effectively regularize the learning process, meanwhile do not reveal clear evidence towards a superior representation space.
The lack of knowledge regularizations may limit the potential for better representations.

To this end, we propose incorporating the structural knowledge to regularize the cross-modal representation learning from two aspects:
(1) {\bf Pull together similar videos with knowledge objective.}
We regularize the video representation learning through a structural knowledge prediction objective, which essentially attracts videos of similar semantics together in the representation space.
(2) {\bf Push apart cross-modal hard negatives with knowledge guidance.}
We augment the cross-modal contrastive objective with a knowledge-guided hard negative sampler, so that the model can efficiently learn desired semantics with subtle variations.

Specifically, we propose a Cross-modaL knOwledge-enhanced Pre-training (CLOP) method with Knowledge Regularization, consisting of two key components as follows:
\begin{enumerate}
\item A simple yet effective {\bf Structural Knowledge Prediction (SKP) Proxy Task}.
  It first represents multi-modal content as a generalized form of ``structural knowledge''.
  Then, a pre-train task is hereby proposed to model video representations with the exploited ``structural knowledge'' as pseudo-labels.
  Essentially, the pseudo-labels act as learnable semantic centers in the representation space.
\item A novel sampling approach for {\bf Knowledge-Guided Contrastive Learning (KCL) with Hard Negatives}.
  This module augments the contrastive pre-train task by a light-weight multi-level hard negative sampling method:
  a) The instance-level sampling retrieves mutually similar samples to form batches of hard negatives;
  b) Based on that, the batch-level sampling selects eligible batches with the aid of the aforementioned knowledge.
\end{enumerate}

We evaluate our method on four standard text-video retrieval tasks and one multi-choice QA task.
Our experiments show clear improvements in performance across all tasks and datasets considered, outperforming prior works by a substantial margin.
Besides, we provide detailed ablation analysis and insights of how our methods affect the latent representation space in the view of \textit{alignment} and \textit{uniformity}~\cite{wang2020understanding}.
The analysis demonstrates that our knowledge regularizations promote the \textit{uniformity} property to help data points better ``spreading'' on the representation space.
It results in better separability of the representations and benefits the cross-modal retrieval tasks.

To summarize, we make the following contributions:
\begin{enumerate}
\item
  We propose a video-text pre-training paradigm to improve the representation learning with knowledge regularizations, which consists of a novel SKP prediction task and a KCL hard negative sampling approach.
\item
  We demonstrate clear improvements of our methods on five downstream tasks, achieving state-of-the-art performance compared to prior works.
\item
  We conduct in-depth analysis of the representation distributions of our model and provide intuitive explanations to the mechanism of our method.
\end{enumerate}


\section{Related Work}
\label{sec:related}

\begin{figure*}[ht]
  \begin{center}
  	\includegraphics[width=1.0\linewidth]{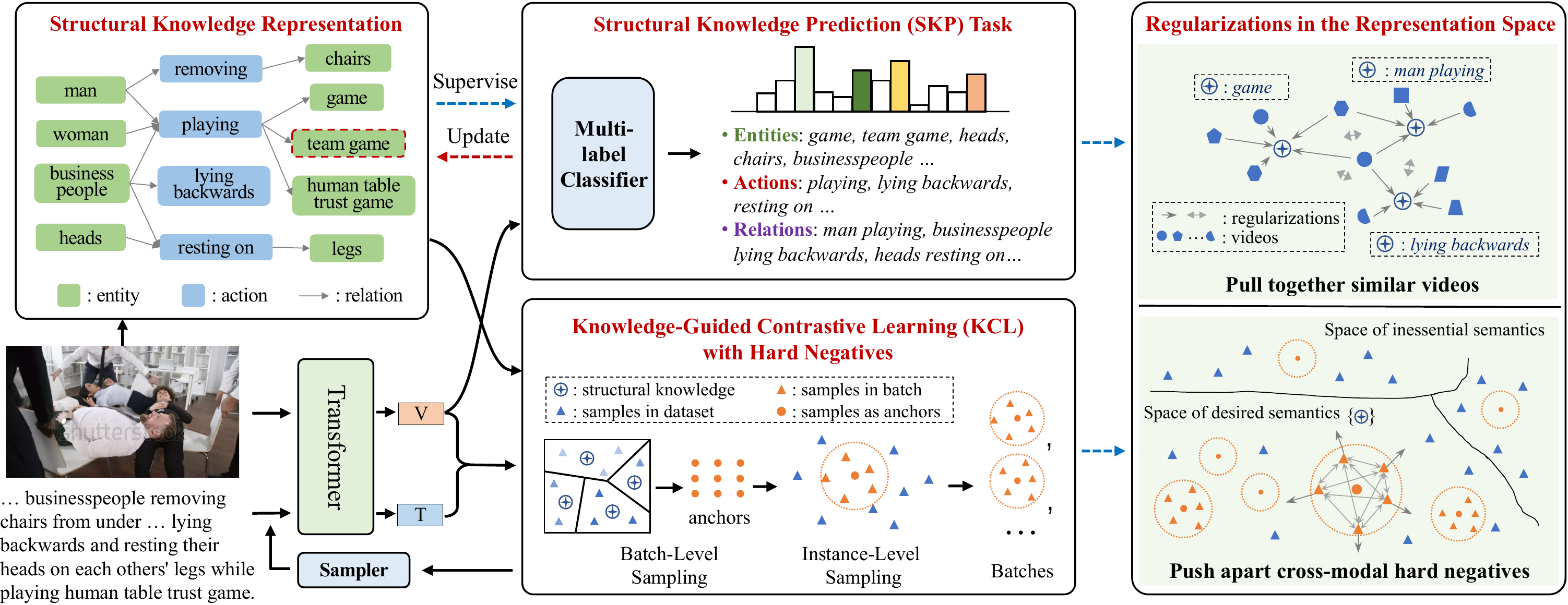}
  \end{center}
  \caption{{\bf Overview of our proposed method.}}
\label{fig:overview}
\end{figure*}

\subsection{Video-and-Language Pre-training}

Since the release of several large-scale video-text datasets~\cite{miech2019howto100m,bain2021frozen},
there have been a lot of works leveraging these corpora for video-text pre-training, including alleviating data noise~\cite{miech2020end,xu2021videoclip}, designing better model architectures~\cite{li2020hero,zhu2020actbert,luo2020univl,xu2021vlm} or proposing new objectives~\cite{li2020hero,patrick2020support}.

Briefly, the dominant paradigm of video-language pre-training is to train powerful transformer-based encoders with various self-supervised proxy tasks.
The canonical pre-train objectives include conditioned masked modeling and cross-modal video-text matching~\cite{chen2019uniter,li2020unicoder}.
As for the video-text matching, the existing works usually leverage cross-modal contrastive objectives to learn the semantic alignments.
Besides, there are other works modeling frame orders~\cite{li2020hero} or using generative objectives~\cite{patrick2020support}.

Beyond the existing works, we propose a novel method to incorporate knowledge regularizations for video-text pre-training, and show its effectiveness through abundant experiments and insights.

\subsection{Incorporating Structural Knowledge}

The knowledge-enhanced pre-training methods first emerge in the NLP fields.
The researchers incorporate structural knowledge in different aspects, such as augmenting raw textual data~\cite{liu2020k}, integrating knowledge embeddings~\cite{zhang2019ernie,liu2021kg}, or constraining masking objectives~\cite{sun2019ernie}.

For the video-text scenario, there are pioneering works trying to incorporate object-level structural knowledge into pre-training.
More recently, OA-Trans~\cite{wang2021object} explicitly utilizes the object tags and bounding boxes in the training process.
ALPRO~\cite{li2021align} proposes to obtain visually-grounded entities by a standalone prompter without requiring off-the-shelf object detectors.
However, most of the existing works are limited to use only object-level information as model inputs, ignoring the structural relationships (e.g., subject-predicate knowledge, etc.).

In contrast, our work expands the symbolic concepts (e.g., objects, relations, etc.) to a generalized form of ``structural knowledge'' by integrating various structural relationships.

\subsection{Contrastive Learning with Hard Negatives}

The key idea behind contrastive learning~\cite{le2020contrastive} is to learn a general feature function by discriminating positive and negative samples -- making the positive pairs attracted and the negative pairs separated~\cite{gutmann2010noise,van2018representation,schroff2015facenet}.
This learning paradiam has been proved very successful in repensenting visual data~\cite{chen2020simple,he2020momentum}, natural languages~\cite{logeswaran2018efficient}, graphs~\cite{sun2019infograph,hassani2020contrastive}, etc.

Hard negatives have been proved essential for contrastive learning systems~\cite{xuan2020hard,cai2020all}.
In the computer vision domain, related works propose various methods to explore negatives,
such as maintaining a memory queue to enrich negatives~\cite{he2020momentum}, designing sampling distributions~\cite{robinson2020contrastive} or generating negatives in latent space~\cite{kalantidis2020hard}.
UNIMO~\cite{li2020unimo} constructs hard textual negatives by re-writing sentences for image-text pre-training.
Recently, several works propose to retrieve nearest neighbors as contrastive negatives for dense text retrieval~\cite{xiong2020approximate} or video-text pre-training~\cite{xu2021videoclip}.

In this work, we propose a knowledge-guided approach for contrastive learning with hard negatives (KCL).
Our method designs a multi-level sampling method to realize an instance-level hard negative sampling and a controllable batch-level sampling to emphasize desired semantics.

\section{Our method}
\label{sec:model}

In this work, we propose a Cross-modaL knOwledge-enhanced Pre-training (CLOP) method.
The overall framework is illustrated in Fig.~\ref{fig:overview}.

\subsection{Architecture Overview}

We use transformer~\cite{vaswani2017attention} as the backbone of our model, and preserve several canonical pre-train tasks.
Specifically, we follow the architecture of HERO~\cite{li2020hero} to establish a base pre-training framework.
Our base model mainly consists of a shared multi-modal encoder for representing texts and videos, four proxy tasks including Masked Language Modeling (MLM), Masked Frame Modeling (MFM), Frame Order Modeling (FOM) and Video-Text Matching (VTM).
The VTM task learns from contrastive samples using hinge-based triplet loss.
Considering the cost of data loading and the effeciency of computing, we follow~\cite{li2021value} to pre-extract 2D + 3D video features~\cite{radford2021learning,feichtenhofer2019slowfast} as input, instead of feeding the model with raw video as several recent end-to-end arts~\cite{lei2021less,bain2021frozen} do.

As shown in Fig.~\ref{fig:overview}, we illustrate the overview of our method.
The video and text pairs are fed into transformer-based encoders to obtain latent representations.
We exploit the cues of the text and video to build an explicit form of ``structural knowledge representation'', including entities, actions, and relations.
Based on that, we design two knowledge regularization components. 

The SKP proxy task pulls together the videos of similar semantics with multi-label classification, where the ``structural knowledge'' serves as pseudo-labels to supervise the learning process (Sec.~\ref{sec:model_skp}).
As the SKP module learns a general understanding of videos, it can update and complement the ``structural knowledge'' with its predicted results (marked with red dashed line boxes in Fig.~\ref{fig:overview}) when it converges.
The KCL module performs sampling for contrastive learning to push apart the cross-modal hard negatives under knowledge guidance (Sec.~\ref{sec:model_cl}).
During the batch-level sampling, the aforementioned knowledge specifies a partition of semantic spaces and guides the model to select proper batch anchors to emphasize the learning of desired semantics.
According to the anchors, the instance-level sampling procedure assembles batches of hard negatives (mutually similar samples) for contrastive learning.
Altogether, these two knowledge regularizations collaborate together to form a superior cross-modal representation space.

\subsection{Structural Knowledge Prediction as Proxy Task}
\label{sec:model_skp}

This module represents multi-modal data as a generalized symbolic form of ``structural knowledge''.
Meanwhile, a cross-modal prediction task is hereby proposed to model video representations with the exploited ``structural knowledge''.

\subsubsection{Structural Knowledge Representations}

We represent multi-modal raw data as symbolic concepts, then expand these concepts by incorporating ``structural knowledge'', i.e., exploiting the structural relationships (e.g., subject-predicate knowledge, superordinate concept knowledge, etc.) to build a generalized form of representation.

For a sample $e=(v, t)$, we first collect two types of symbolic concepts from the textual and visual inputs.
In detail, for the textual side, we utilize the Stanford CoreNLP Toolkit~\cite{manning2014stanford} and Stanford Part-Of-Speech Tagger~\cite{toutanova2003feature} to extract all meaningful concepts (e.g., nouns and verbs) from the text.
For the visual side, we generate scene graphs from videos with ~\cite{han2021image} and remain the names of recognized objects as concepts.
Altogether, we get a list of atom concepts $\{ c_{i} \}=\{ c_{0}, c_{1}, \cdots \}$ for each sample.

Afterward, we exploit various structural relationships to expand the atom concepts.
Generally, we denote each structural relationship as an operation $\mathcal{G}_{rel}(\cdot)$ to transform a set of concepts into a new set of concepts.
For example, the subject-predicate (SP) relationship is important for discriminating fine-grained semantics when there are multiple subjects (agents) but a single predicate (action).
we incorporate this kind of knowledge by create another type of concepts (subject-predicate phrases) by combining each predicate with its corresponding subject, i.e., $\mathcal{G}_{sp}(\{ c_{i}\})=\{ c^{sp}_{0}, c^{sp}_{1}, \cdots \}$.
We also use the as-is relationship $\mathcal{G}_{as}(\{ c_{i}\})=\{ c_{i} \}$ to remain the atom concepts.
More structural relationships could be considered to further enhance the representation, such as linking them with outside concepts based on external KBs, etc.
In the end, we build the ``structural knowledge'' representations by merging the output of every transform operation: $\boldsymbol{o}=\{o_{i}\}=\bigcup\mathcal{G}_{rel}(\{c_{i}\})$, $rel \in \{ sp, as, \cdots \}$.

Overall, the ``structural knowledge'' representations hold the following benefits:
\begin{enumerate}
\item Universal. Representations of different semantic granularity (e.g., entities, actions, subject-predicate relations, etc.) are unified in the flexible form of explicit ``structural knowledge''.
\item Extendable. It is possible to link the multi-modal content with outside structural resources (e.g., superordinate concept), thus expanding the realm of knowledge.
\end{enumerate}

\subsubsection{Structural Knowledge Prediction Proxy Task}

We regard the aforementioned structural knowledge as pseudo-labels, and propose a simple yet effective structural knowledge prediction (SKP) proxy task to regularize the representation of videos.

Given the ``structural knowledge'' representation $\boldsymbol{o}_{e}$ for each sample $e=(v, t)$ in the pre-train dataset $\mathcal{D}=\{e_{0}, e_{1}, \cdots,e_{|\mathcal{D}|-1}\}$,
we first collect the most freqently (e.g., top 15000) appeared concepts across the pre-train dataset, i.e., $\boldsymbol{\mathcal{O}}=\mathrm{topK}(\{ o | o \in \boldsymbol{o}_{e_{i}}, \forall e_{i} \in \mathcal{D} \})$.
Then, each video $v$ is fed into the encoder $f(\cdot)$ and a Multi-Layer Perceptron (MLP) to predict the probability of ``structural knowledge'' targets: $P(\boldsymbol{y}|v;\theta)=\sigma(\mathrm{MLP}(f(v)))$.

The SKP task is formulated as a multi-label classification with binary cross-entropy loss:
{
  \small
\begin{align}
  &\mathcal{L}^{\mathsf{(SKP)}} = -\sum_{i} \left[ y^{*}_{i} \log P(y_{i}|v;\theta) + (1 - y^{*}_{i}) \log (1 - P(y_{i}|v;\theta)) \right],
\end{align}
} where $y^{*}_{i}$ is the ground-truth label indicating whether concept $\mathcal{O}_{i}$ exists in $\boldsymbol{o}_{e}$, $\theta$ represents the neural network parameters.

Note that, our SKP task is obviously different from existing pre-train tasks in several aspects:
\begin{enumerate}
\item Compared with other prediction tasks, e.g., Masked Language Modeling (MLM)~\cite{devlin2018bert}, 
  we learn visual representations with cross-modal cues, instead of modeling contextual language.
  Besides, the prediction targets are beyond the word vocabulary.
\item Compared with other cross-modal tasks, e.g., visual-text matching~\cite{lu2019vilbert},
  we construct flexible prediction targets to learn semantic alignments with multiple granularities.
\end{enumerate}

As a proxy pre-train task, the SKP task regularizes the embedding space by pulling together the videos of similar semantics, where the ``structural knowledge'' specifies learnable semantic centers.
A detailed analysis of how the SKP task affects the embedding space is provided in the experiments section (Sec.~\ref{sec:exp_insights}).

\subsection{Knowledge-Guided Contrastive Learning with Hard Negatives}
\label{sec:model_cl}

We propose a knowledge-guided sampling approach for contrastive learning (KCL) with hard negatives.
It is a two-stage sampling procedure:
1) For the batch-level, the model is guided to select proper anchor samples for every batch to emphasize the learning of desired semantics.
2) For the instance-level, according to the batch anchors, we assemble batches of hard negatives (mutually similar samples) from the whole training dataset for contrastive learning.

We first introduce the instance-level hard negative sampling for contrastive learning as a basis, then elaborate on the knowledge-guided batch-level sampling.

\subsubsection{Contrastive Learning with Instance-Level Hard Negative Sampling}
\label{sec:model_cl_base}

The hard negative samples are more informative than the easy ones for learning a good embedding space~\cite{xuan2020hard,cai2020all}.
Inspired by recent works~\cite{xiong2020approximate,xu2021videoclip}, we implement a custom sampling procedure to gather batches of hard negatives from the training set, by retrieving clusters of mutually similar samples.

The similarity of two individual samples is measured based on their latent representations in a common embedding space.
For each sample $e=(v,t)$, its embedding is calculated as the average of its visual and textual representations.
We adopt a light-weight ``self-distilled, online-updating'' approach to maintain the embedding of each sample:
\begin{enumerate}
\item Self-distilled. The embeddings are computed by the model itself without requiring any extra resources;
\item Online-updating. We update the embeddings on-the-fly during the previous training epoch, thus introducing no computational overhead.
\end{enumerate}

At the start of each training epoch, we first sample $L$ (the number of batches per epoch) samples from the training set, in which each sample is regarded as the representative anchor of each batch.
Then, we collect each batch of samples according to the anchor by retrieving its $k$ (a number larger than the batch size) nearest neighbors~\footnote{We use faiss~\cite{johnson2019billion} for efficient k-nearest neighbor searching.} from the training set and randomly choose $N$ (the batch size) neighboring samples to finally form a batch.
In this way, we gather $L$ hard negative batches, where each sample is a hard negative of each other in the same batch.

Note that during the sampling procedure, a part of training samples may appear zero or multiple times in one epoch.
To ensure that every sample embedding is available for nearest neighbor searching in the next epoch, we organize the missing training samples into random batches $\mathcal{R}$.
Combined with the hard negative batches $\mathcal{C}$, we randomly shuffle the order of the $\mathcal{C}+\mathcal{R}$ batches for this epoch.
The detailed sampling procedure is described in Algorithm~\ref{alg:sampling}.

\begin{algorithm}
  \renewcommand{\algorithmicrequire}{\textbf{Input:}}
  \renewcommand{\algorithmicensure}{\textbf{Output:}}
  \caption{Training with KCL Hard Negative Sampling.}
  \label{alg:sampling}
  \small
  \begin{algorithmic}[1]
    \REQUIRE 1) all samples $\mathcal{D}=\{e_{0}, e_{1}, \cdots,e_{|\mathcal{D}|-1}\}$, where $e_{i}=(v_{i},t_{i})$; \\
    2) the hyper-parameter batch-size $N$; \\ 3) the encoder $f(\cdot)$.
    \ENSURE A list of batches $\mathcal{B}=\{ B_{0}, B_{1}, \cdots, B_{|\mathcal{B}|-1}\}$, \\
    where each $B_{i}$ consists of $N$ samples: $\{ e_{b_{i,0}}, e_{b_{i,1}}, \cdots, e_{b_{i,N-1}}\}$ 
    \STATE $\mathcal{M}$ = an empty embedding memory cache.
    \FORALL{epoch}
    \IF{$\mathcal{M}$ is not empty}
    \STATE \% {\it batch-level sampling}
    \STATE Draw $L$ anchors $\mathcal{A}=\{e_{a_{0}}, \cdots, e_{a_{L-1}}\}$ from $\mathcal{D}$ as in Sec.~\ref{sec:batch_level}
    \STATE \% {\it instance-level sampling}
    \FORALL{anchor $i \in \{0, \cdots, L-1\}$}
    \STATE Retrieve a cluster of samples using nearest neighbor searching: \\
    \quad $\mathcal{C}_{i} \sim k\mathrm{NN}(\mathcal{M}[e_{a_{i}}],k=2N;\mathcal{M}[\cdot])$, where $|\mathcal{C}_{i}|=N$
    \ENDFOR
    \STATE \% {\it add the non-visited samples}
    \STATE $\mathcal{R}=\mathrm{Random\_Assemble\_Batches}(\mathcal{D} - \{e | e\in\mathcal{C}_{i}, 0 <= i < L \})$
    \STATE $\mathcal{B} = \mathrm{Shuffled}(\mathcal{C} + \mathcal{R})$
    \ELSE
    \STATE $\mathcal{B} = \mathrm{Random\_Assemble\_Batches}(\mathcal{D})$
    \ENDIF
    \FORALL{batch $B_{i} \in \mathcal{B}$}
    \STATE $\forall j \in \{0, \cdots, N-1\}$
    \STATE Forward: infer embeddings $\{ (z^{v}_{b_{i,j}},z^{t}_{b_{i,j}}) \}=f(\{ e_{b_{i,j}}\})$.
    \STATE Update $\mathcal{M}[e_{b_{i,j}}]=(z^{v}_{b_{i,j}} + z^{t}_{b_{i,j}}) / 2$
    \STATE Backward: update the encoder parameters $f(\cdot)$.
    \ENDFOR
    \STATE Build $k\mathrm{NN}$ retriever index with $\mathcal{M}$.
    \ENDFOR
  \end{algorithmic}
\end{algorithm}

In the training stage, the textual and visual input are fed into the encoder $f(\cdot)$ to obtain the normalized representations $\{ (z^{v}_{i},z^{t}_{i}) \}=f(\{e_{i}\})$.
For a batch of features $\{(z^{v}_{0},z^{t}_{0}),(z^{v}_{1},z^{t}_{1}),\cdots,(z^{v}_{N-1},z^{t}_{N-1})\}$, every sample is a contrastive negative of the others.
We compute the symmetrical hinge-based triplet loss for a batch as follows:

\begin{align}
  &\mathcal{L}^{\mathsf{(t2v)}}_{i,j} = \mathrm{max}(0, \delta + sim(z^{t}_{i}, z^{v}_{j}) - sim(z^{t}_{i}, z^{v}_{i})), \\
  &\mathcal{L}^{\mathsf{(v2t)}}_{i,j} = \mathrm{max}(0, \delta + sim(z^{v}_{i}, z^{t}_{j}) - sim(z^{v}_{i}, z^{t}_{i})), \\
  &\mathcal{L}^{\mathsf{(KCL)}}=\frac{1}{N}\sum_{i}\sum_{j \neq i}(\mathcal{L}^{t2v}_{i,j} + \mathcal{L}^{v2t}_{i,j}), \quad \forall i,j \in \{0, \cdots, N-1\}
\end{align}
where $\delta$ is the hyper-parameter margin, $sim$ operator computes the cosine similarity of two normalized features.

\begin{figure}[h]
  \centering
  \includegraphics[width=.45\textwidth]{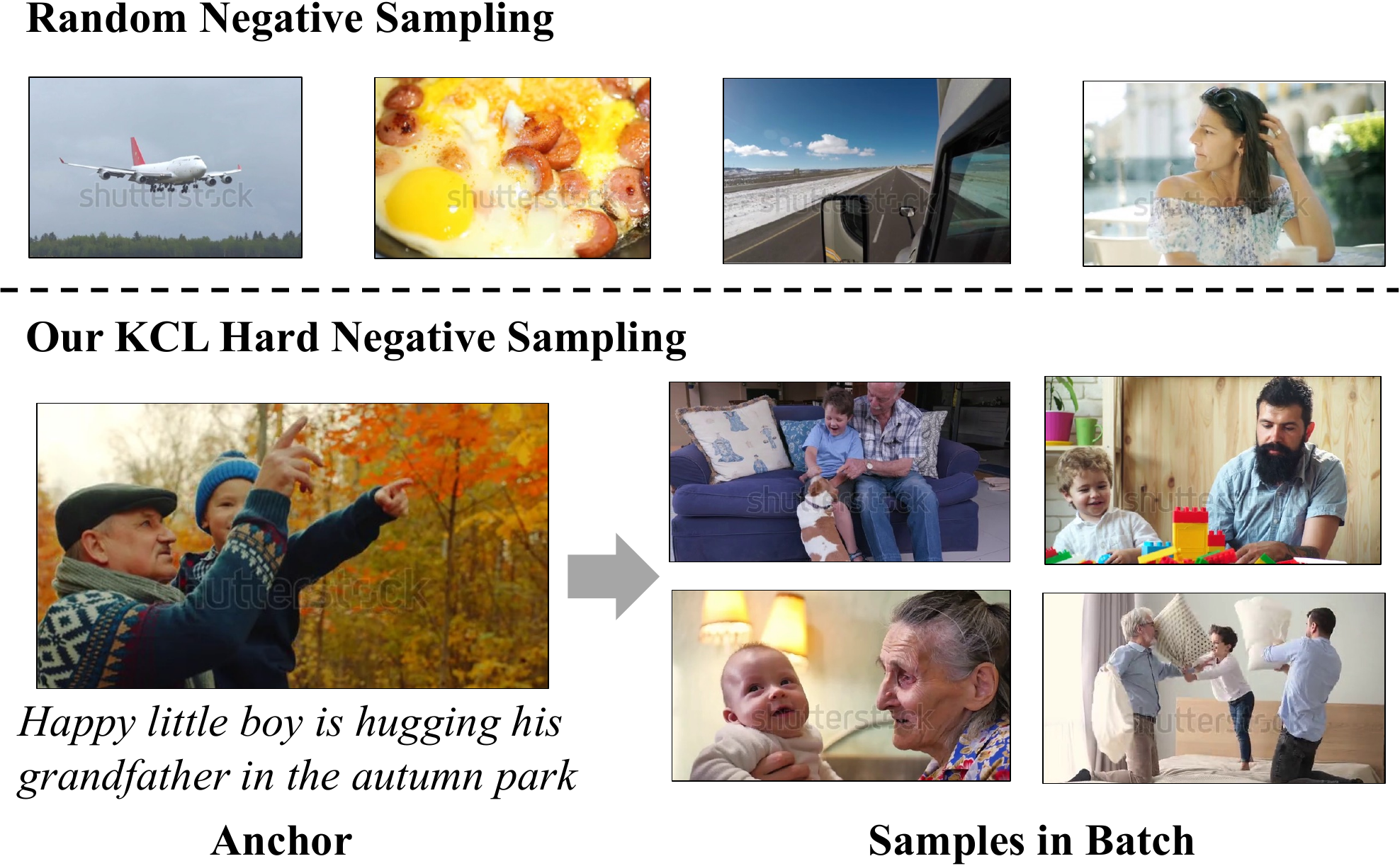}
  \caption{Comparisons of the random negative sampling and our KCL hard negative sampling.}
  \label{fig:sampling}
\end{figure}

\subsubsection{Knowledge-Guided Batch-Level Sampling}
\label{sec:batch_level}

In the previous section, we have developed an instance-level hard negative sampling method for gathering samples in a batch.
It constructs each batch with mutually similar samples, in other words, each batch is assembled around a specific ``topic''.

However, we notice that there is a ``distribution discrepancy'' issue for the sampling procedure:
\begin{enumerate}
\item Redundancy. A considerable number of ``topics'' are frequently sampled during pre-training but are usually unimportant for video understanding, e.g., abstract scenes, boring landscape views, etc.;
\item Insufficiency. In contrast with redundancy, there are numerous ``topics'' very important for video understanding but not sufficiently learned, e.g., human actions, events, etc.
\end{enumerate}

Based on these observations, we propose a knowledge-guided batch-level sampling component to address the ``distribution discrepancy'' issue.
This component augments the instance-level sampling algorithm of Sec.~\ref{sec:model_cl_base} as shown in Alg.~\ref{alg:sampling} line 5.
It modulates the sampling distribution of batch anchors to emphasize the desired but insufficient learning ``topics'', and suppress the inessential but redundant learning ``topics''.
In this procedure, we use the prediction results of the aforementioned SKP task (Sec.~\ref{sec:model_skp}) to identify the ``topics'' of each batch.
In this sense, we denote the specified collections of ``topics'' (desired semantics) as knowledge to guide the batch sampling and model learning.

In detail, we adopt a heuristic approach to emphasize the human-action-related semantics.
When sampling batch anchors, we increase the sampling probabilities for the anchors that involve human actions, at the same time preventing the selection of irrelevant anchors.
We validate the feasibility of our instance-level and batch-level sampling methods against random negative sampling for contrastive learning in the ablations (Sec.~\ref{sec:exp_ablations}).
A qualitative visualization of our KCL sampling are shown in Fig.~\ref{fig:sampling}.

~\\
We close the section by pointing out that, our SKP (Sec.~\ref{sec:model_skp}) and KCL (Sec.~\ref{sec:model_cl}) knowledge regularizations collaborate together to encourage a superior cross-modal representation space (Sec.~\ref{sec:exp_insights}).

\begin{table*}[h]
  \centering
  \caption{Experiments of text-to-video retrieval on 1k-A MSRVTT test set.
  } 
  \scalebox{1.0}{
    \begin{tabular}{l|llccccc}
      \toprule
      Method & Years & Pretrain Dataset & AveR & R@1 & R@5 & R@10 & MedR\\
      \midrule
      ActBERT~\cite{zhu2020actbert} & CVPR'20 & [136M] HowTo100M & 38.7 & 16.3 & 42.8 & 56.9 & 10.0\\
      UniVL~\cite{luo2020univl} & Arxiv'20 & [136M] HowTo100M & 44.6 & 21.2 & 49.6 & 63.1 & 6.0\\
      MMT~\cite{gabeur2020multi} & ECCV'20 & [136M] HowTo100M & 51.1 & 26.6 & 57.1 & 69.6 & 4.0\\
      SupportSet~\cite{patrick2020support} & ICLR'21 & [136M] HowTo100M & 52.6 & 30.1 & 58.5 & 69.3 & 3.0\\
      ClipBERT~\cite{lei2021less} & ICCV'21 & [5.6M] COCO, VisGenome & 42.9 & 22.0 & 46.8 & 59.9 & 6.0\\
      VideoCLIP~\cite{xu2021videoclip} & EMNLP'21 & [136M] HowTo100M & 51.0 & 30.9 & 55.4 & 66.8 & -\\
      Frozen~\cite{bain2021frozen} & ICCV'21 & [5.5M] WebVid2M, CC3M  & 53.7 & 31.0 & 59.5 & 70.5 & 3.0\\
      ALPRO~\cite{li2021align} & Arxiv'21 & [5.5M] WebVid2M, CC3M & 55.9 & 33.9 & 60.7 & 73.2 & 3.0 \\
      OA-Trans~\cite{wang2021object} & Arxiv'21 & [2.5M] WebVid2M  & 55.4 & 32.7 & 60.9 & 72.5 & 3.0 \\
      {\bf [ours] base} & - & [2.5M] WebVid2M & 55.4 & 31.7 & 61.6 & 72.8 & 3.0 \\ 
      {\bf [ours] full} & - & [2.5M] WebVid2M & {\bf 57.8} & {\bf 34.0} & {\bf 64.3} & {\bf 75.0} & 3.0 \\
      \midrule
      {\it - zeroshot -} \\
      Frozen~\cite{bain2021frozen} & ICCV'21 & [5.5M] WebVid2M, CC3M & 36.6 & 18.7 & 39.5 & 51.6 & 10.0\\
      {\bf [ours] base} & - & [2.5M] WebVid2M & 36.2 & 18.7 & 39.8 & 50.0 & 10.0 \\ 
      {\bf [ours] full} & - & [2.5M] WebVid2M & {\bf 38.7} & {\bf 21.3} & {\bf 42.4} & {\bf 52.5} & {\bf 9.0} \\
      \bottomrule
	\end{tabular}
  }
  \label{tab:msrvtt}
\end{table*}


\section{Experiments}
\label{sec:exp}

\begin{figure*}[ht]
  \begin{center}
  	\includegraphics[width=.85\linewidth]{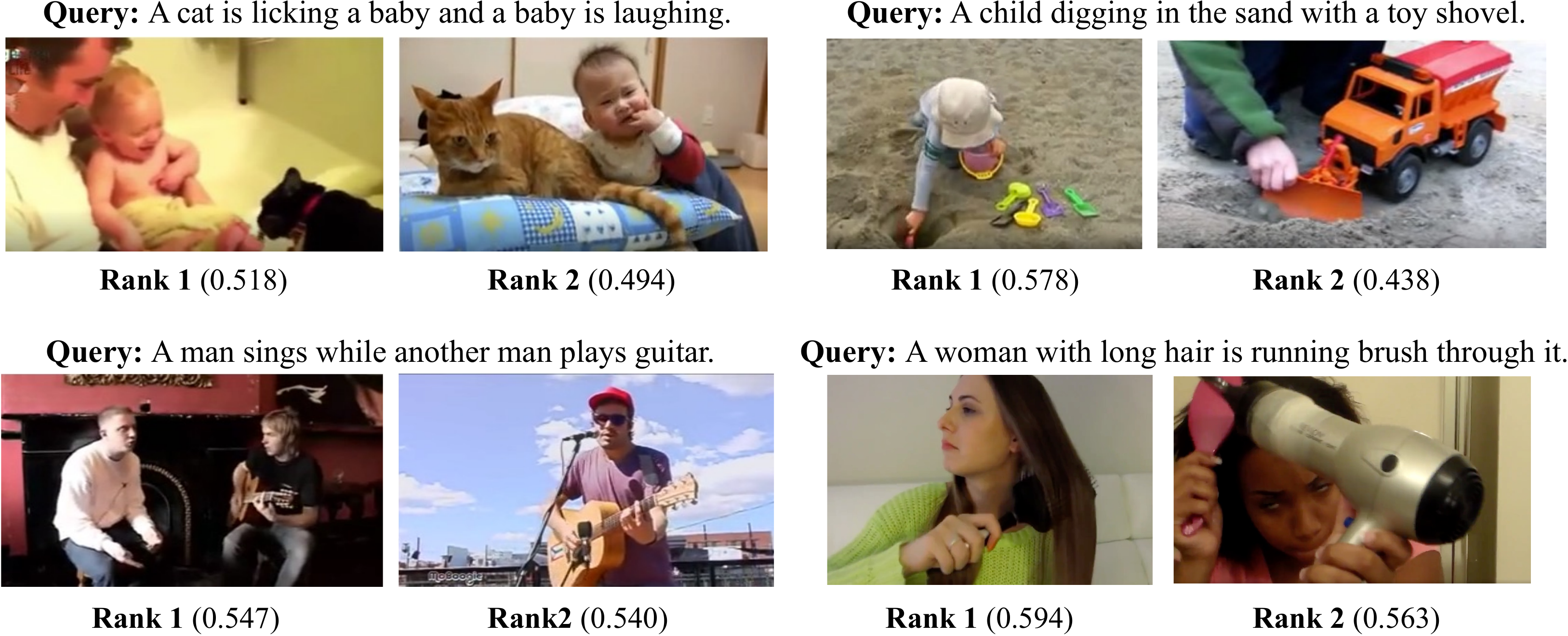}
  \end{center}
  \caption{Visualized examples on MSRVTT and VATEX text-to-video retrieval. We provide the Rank 1 and Rank 2 retrieved videos and their query-video cosine similarity scores. Our method can distinguish the subtle semantic differences.}
  \label{fig:visualization}
\end{figure*}

In this section, we first describe the overall experiment settings in Sec.~\ref{sec:exp_setting}, then present the main results on several end tasks in Sec.~\ref{sec:exp_sota}.
We also provide in-depth analysis of our method through ablations studies (Sec.~\ref{sec:exp_ablations}), insights of the embedding space (Sec.~\ref{sec:exp_insights}) and visualized examples (Fig.~\ref{fig:visualization}).

\subsection{Experiment Settings}
\label{sec:exp_setting}

Our experiment follows the conventional ``pretrain + finetune'' two-stage paradigm.
We pre-train our models without/with our knowledge regularization techniques (i.e., base/full variants of our method),
then finetune on downstream tasks with the pre-trained weights as model initialization.

Our method only takes effect in the pre-train stage and not affects the finetune stage.
The changes in the resulting finetune scores clearly reflect the effectiveness of our proposed method.
Additionally, we also provide zero-shot results, which should be more straightforward for evaluating pre-training strategies.

\spara{Pre-Training Dataset}
We perform video-text pre-training on the recently released large-scale WebVid2M dataset~\cite{bain2021frozen}.
The WebVid2M dataset consists of 2.5 million video-text pairs scraped from the web.
Despite that the dataset is an order of magnitude smaller than the Howto100M~\cite{miech2019howto100m} dataset, it demonstrates better alignments between text semantics and video content.

\spara{End Tasks}
We evaluate our methods on four retrieval tasks and one multi-choice QA task.

The MSRVTT~\cite{xu2016msr} is a well-known dataset for text-video retrieval.
This dataset consists of 10K videos, where each video is associated with 20 text descriptions.
We follow the official 1k-A split~\cite{yu2018joint} for training and evaluation.

The MSVD~\cite{chen2011collecting} dataset consists of 1,970 videos with about 40 text descriptions per video.
This dataset is split into 1,200 videos for training, 100 videos for validation, and 670 videos for testing.
We report the text-video retrieval results on the standard 670 test videos.

The VATEX ~\cite{wang2019vatex} dataset is a large-scale, high-quality multilingual video-text dataset.
This dataset contains 34,991 videos, of which 25,991 videos are for training, 3K for validation, and 6K for testing.
Since the test annotations are not available, we follow the widely-used HGR's~\cite{chen2020fine} protocol which splits the original validation set into 1,500 test videos and 1,500 validation videos.
We only use the English annotations for fair comparisons.

The DiDeMo ~\cite{anne2017localizing} dataset contains 10K flickr videos and 40K localized text annotations.
We follow ~\cite{liu2019use,lei2021less,bain2021frozen} to evaluate paragraph-to-video retrieval task,
where text annotations of the same video are concatenated into a single query.
We report the results without the ground-truth localization proposals (as described in ~\cite{bain2021frozen}).

The MSRVTT multi-choice test~\cite{yu2018joint} is formulated as a multi-choice QA task, which requires the model to select the correct (most relevant) answer from 5 textual candidates based on a given video.
This task shares the MSRVTT training videos and evaluates video-text relevance similar to retrieval tasks.

\spara{Evaluation Metrics}
We report the Average Recall at rank K ($R@K, K\in\{1,5,10\}$) and the Median Rank ($MedR$) metrics for retrieval tasks.
The higher $R@K$ and the lower $MedR$ indicate better performance.
For the MSRVTT multi-choice test, we report the accuracy ($Acc.$).
Note that, in order to reduce the variance in finetuning experiment, the scores are averaged from three repeated experiments with casually selected random seeds.

\begin{table}[h]
  \centering
  \caption{MSVD test set with 670 videos.}
  \scalebox{0.8}{
    \begin{tabular}{l|ccccc}
      \toprule
      Method & AveR & R@1 & R@5 & R@10 & MedR \\
      \midrule
      SupportSet~\cite{patrick2020support} & 47.2 & 23.0 & 52.8 & 65.8 & 5.0\\
      SupportSet-PT~\cite{patrick2020support} & 53.8 & 28.4 & 60.0 & 72.9 & 4.0\\
      Frozen~\cite{bain2021frozen} & 58.2 & 33.7 & 64.7 & 76.3 & 3.0\\
      {\bf [ours] base} & 58.5 & 33.2 & 65.5 & 76.9 & 3.0 \\
      {\bf [ours] full} & {\bf 61.5} & {\bf 37.5} & {\bf 67.9} & {\bf 79.0} & {\bf 2.0} \\
      \midrule
      {\it - zeroshot -} \\
      {\bf [ours] base} & 52.0 & 27.9 & 58.0 & 70.2 & 4.0 \\
      {\bf [ours] full} & 55.9 & 33.5 & 61.7 & 72.5 & 3.0 \\
      \bottomrule
    \end{tabular}
  }
  \label{tab:msvd}
\end{table}

\spara{Implementation Details}
We pre-train a 6-layer 12-head transformer (initialized with Roberta~\cite{liu2019roberta} weights) as the video and text encoder with five proxy tasks (SKP and KCL, plus other three tasks following~\cite{li2020hero}).
The multiple pre-train tasks are switched at random for each batch.
We use 4 A100 GPUs and pre-train for about 6 days, with batch size 128 per GPU and learning rate 2e-5.
Please refer to the Supplementary Material for more implementation details.

\subsection{Overall Results on End Tasks}
\label{sec:exp_sota}

The overall results on five downstream tasks are listed in Tab.~\ref{tab:msrvtt}-\ref{tab:msrvtt-mc}.
We provide two model variants of ours in these tables.
As the name indicates, the {\bf [ours] base} variant represents the base model we built on top of, while the {\bf [ours] full} represents the full model with our knowledge regularization techniques (SKP + KCL).
The direct comparisons between the two variants can provide convincing evidence for assessing the effects of our proposed method.

Compared with the base model, we obtain on average $+2.9\%$ finetune and $+3.2\%$ zero-shot $\mathrm{AveR}@\{1,5,10\}$ boosts across the MSRVTT (Tab.~\ref{tab:msrvtt}), MSVD (Tab.~\ref{tab:msvd}), VATEX (Tab.~\ref{tab:vatex}) and DiDeMo (Tab.~\ref{tab:didemo}) retrieval tasks.
Besides retrieval, we also conduct a quick evaluation of the MSRVTT multi-choice test in Tab.~\ref{tab:msrvtt-mc}.
In general, our method shows clear improvements across all the tasks above and outperforms the state-of-the-art methods by a substantial margin.
The overall results demonstrate the effectiveness of our method and prove it generalizes well across various datasets.

\begin{table}[h]
  \centering
  \caption{VATEX HGR-1k5 test set.}
  \scalebox{0.8}{
    \begin{tabular}{l|ccccc}
      \toprule
      Method & AveR & R@1 & R@5 & R@10 & MedR\\
      \midrule
      HGR~\cite{chen2020fine} & 64.0 & 35.1 & 73.5 & 83.5 & 2.0\\
      SupportSet~\cite{patrick2020support} & 72.0 & 44.6 & 81.8 & 89.5 & 1.0\\
      SupportSet-PT~\cite{patrick2020support} & 72.9 & 45.9 & 82.4 & 90.4 & 1.0\\
      {\bf [ours] base} & 78.1 & 54.8 & 87.0 & 92.6 & 1.0 \\
      {\bf [ours] full} & {\bf 79.7} & {\bf 56.6} & {\bf 88.6} & {\bf 93.9} & 1.0 \\
      \midrule
      {\it - zeroshot -} \\
      {\bf [ours] base} & 33.3 & 19.5 & 37.2 & 43.1 & 32.0 \\
      {\bf [ours] full} & 35.8 & 22.8 & 39.5 & 45.1 & 19.0 \\
      \bottomrule
    \end{tabular}
  }
  \label{tab:vatex}
\end{table}

\begin{table}[h]
  \centering
  \caption{DiDeMo test set.}
  \scalebox{0.9}{
    \begin{tabular}{l|ccccc}
      \toprule
      Method & AveR & R@1 & R@5 & R@10 & MedR\\
      \midrule
      CE~\cite{liu2019use} & - & 16.1 & 41.1 & - & 8.0 \\
      ClipBert~\cite{lei2021less} & 40.5 & 20.4 & 44.5 & 56.7 & 7.0 \\
      Frozen~\cite{bain2021frozen} & 54.4 & 31.0 & 59.8 & 72.4 & 3.0 \\
      {\bf [ours] base} & 51.5 & 27.8 & 57.2 & 69.6 & 4.0 \\
      {\bf [ours] full} & {\bf 56.2} & {\bf 32.7} & {\bf 62.7} & {\bf 73.3} & 3.0 \\
      \midrule
      {\it - zeroshot -} \\
      {\bf [ours] base} & 36.9 & 16.2 & 41.1 & 53.4 & 9.0 \\
      {\bf [ours] full} & 40.8 & 20.2 & 45.5 & 56.7 & 8.0 \\
      \bottomrule
    \end{tabular}
  }
  \label{tab:didemo}
\end{table}

\begin{table}[h]
  \centering
  \caption{MSRVTT Multi-choice QA.}
  \scalebox{0.9}{
    \begin{tabular}{c|c}
      \toprule
      Method & Accuracy\\
      \midrule
      JSFusion~\cite{yu2018joint} & 83.4 \\
      ActBERT~\cite{zhu2020actbert} & 85.7 \\
      ClipBert~\cite{lei2021less} & 88.2 \\
      VideoCLIP~\cite{xu2021videoclip} & 92.1 \\
      Ours & {\bf 95.6} \\
      \bottomrule
    \end{tabular}
  }
  \label{tab:msrvtt-mc}
\end{table}

We notice that, a line of CLIP-based methods~\cite{luo2021clip4clip,fang2021clip2video} have achieved remarkable performance on the retrieval tasks.
However, they use the model architecture and weights from CLIP~\cite{radford2021learning}, which is pre-trained on 400 million image-text pairs (150x larger than WebVid2M~\cite{bain2021frozen}), thus are not suitable to compare in our settings.

\subsection{Ablation Studies}
\label{sec:exp_ablations}

As shown in Tab.~\ref{tab:ablations},
we conduct extensive ablation experiments to study the effects of each component in our method.
Considering the heavy computational cost, unless specified, we use a subset of WebVid data (500K / 2.5M) to pre-train our model in ablation studies.
We provide the average recall $\mathrm{AveR}@\{1,5,10\}$ results on the VATEX HGR-1k5 test set in finetuning and zero-shot settings.
Note that, the finetune results are averaged from three repeated experiments with casually selected random seeds.

\subsubsection{Effects of the Model Components}
As shown in Tab.~\ref{tab:ablations} (1), both the SKP and KCL components obtain obvious improvements over the base model ($+1.52\%$ and $+1.04\%$) for finetuning.
Putting them together achieves even better performance ($+1.99\%$).
As for zero-shot evaluation, it shows less improvement ($+0.18\%$) for SKP but greater improvement ($+2.59\%$) for KCL.
We think the reason is that, KCL directly optimizes the cross-modal relevance similar to the retrieval end task, while the SKP task improves the potential to be exploited with finetuning for superior representations.

\subsubsection{Effects of the Structural Knowledge for SKP Task}
The SKP$[*]$ represents SKP variants with different structural knowledge representations.
$[t]$ only uses the atom concepts from the text; $[t+v]$ uses both the textual and visual atom concepts;
$[t+v+s]$ expands all the atom concepts with structural relationships, e.g., the subject-predicate (SP) knowledge.
As shown in Tab.~\ref{tab:ablations} (2), adding the video-side concepts improves zero-shot performance ($+1.48\%$); incorporating the structural relationships brings extra $+0.37\%$ finetune performance gains.

\subsubsection{Effects of the Strategies for KCL sampling}
The KCL[v1] and KCL[v2] represent the negative sampling in Sec.~\ref{sec:model_cl_base} and Sec.~\ref{sec:batch_level}.
As shown in Tab.~\ref{tab:ablations} (3), the relative $+0.38\%$ and $+1.51\%$ improvements support the effectiveness of the knowledge-guided batch-level sampling.
Overall, our method outperforms the base model by a significant margin ($+2.72\%$ and $+6.68\%$).

\begin{table}[hb]
  \centering
  \caption{Ablation results on the VATEX HGR-1k5 test set.}
  \label{tab:ablations}
  \scalebox{0.9}{
    \begin{tabular}{l|cc|cc}
      \toprule
      \multirow{2}*{Ablations} & \multicolumn{2}{|c|}{Finetune} & \multicolumn{2}{|c}{Zero-shot} \\
      \cline{2-5}
      ~ & \multicolumn{1}{|c|}{AveR} & Boost & \multicolumn{1}{|c|}{AveR} & Boost \\
      \midrule
      - Base model                       & & & & \\
      \qquad- with 2.5M data             & 78.14 & +2.22 & 33.27 & +4.68 \\
      \qquad- with 500K data             & 75.92 & 0.00 & 28.59 & 0.00 \\
      \hline
      1) - Model components              & & & & \\
      \qquad- SKP$[t]$                   & 77.44 & +1.52 & 28.77 & +0.18 \\
      \qquad- KCL[v1]                    & 76.96 & +1.04 & 31.18 & +2.59 \\
      \qquad- KCL[v1] + SKP$[t]$         & 77.91 & +1.99 & 32.50 & +3.91 \\
      \hline
      2) - SKP task                      & & & & \\
      \qquad- KCL[v1] + SKP$[t+v]$       & 77.89 & +1.97 & 33.98 & +5.39 \\
      \qquad- KCL[v1] + SKP$[t+v+s]$     & 78.26 & +2.34 & 33.76 & +5.17 \\
      \hline
      3) - KCL sampling                  & & & & \\
      \qquad- KCL[v2] + SKP$[t+v+s]$     & {\bf 78.64} & +2.72 & {\bf 35.27} & +6.68 \\
      \bottomrule
    \end{tabular}
  }
  \label{tab:ablation}
\end{table}

\subsection{Insights of the Representation Space}
\label{sec:exp_insights}

To get further insights, we conduct a quantitative analysis of how our methods affect the latent representation space in the view of \textit{alignment} and \textit{uniformity}~\cite{wang2020understanding}.

The {\it alignment} metric measures the ``closeness'' of positive pairs (i.e., similar samples should have similar features).
We make slight modifications to adapt the formulations to the cross-modal scenario, as the expected distance between positive video-text pairs:
\begin{align}
  \mathcal{L}_{\mathsf{align}}(f;\alpha) \triangleq
  \underset{(v, t) \sim p_{\mathsf{pos}}}{\mathbb{E}} \lVert f(v) - f(t) \rVert_{2}^{\alpha}, \quad \alpha > 0.
\end{align}

The {\it uniformity} metric indicates the ``spreading'' degree of the data points on the hypersphere.
We measure this property for text and videos, respectively.
It is defined as the logarithm of the average pairwise Gaussian potential:
\begin{align}
  \mathcal{L}^{\mathsf{(m)}}_{\mathsf{unif}}(f;\beta) \triangleq
  \log \underset{(x, y) \sim p_{\mathsf{m}}}{\mathbb{E}} [e^{-\beta \lVert f(x) - f(y) \rVert_2^2}], \quad \beta > 0.
\end{align}
where $\mathsf{m} \in \{ \mathsf{vis}, \mathsf{txt} \}$.

\begin{figure}[h]
  \centering
  \begin{tabular}{cccc}
    \includegraphics[width=.45\textwidth]{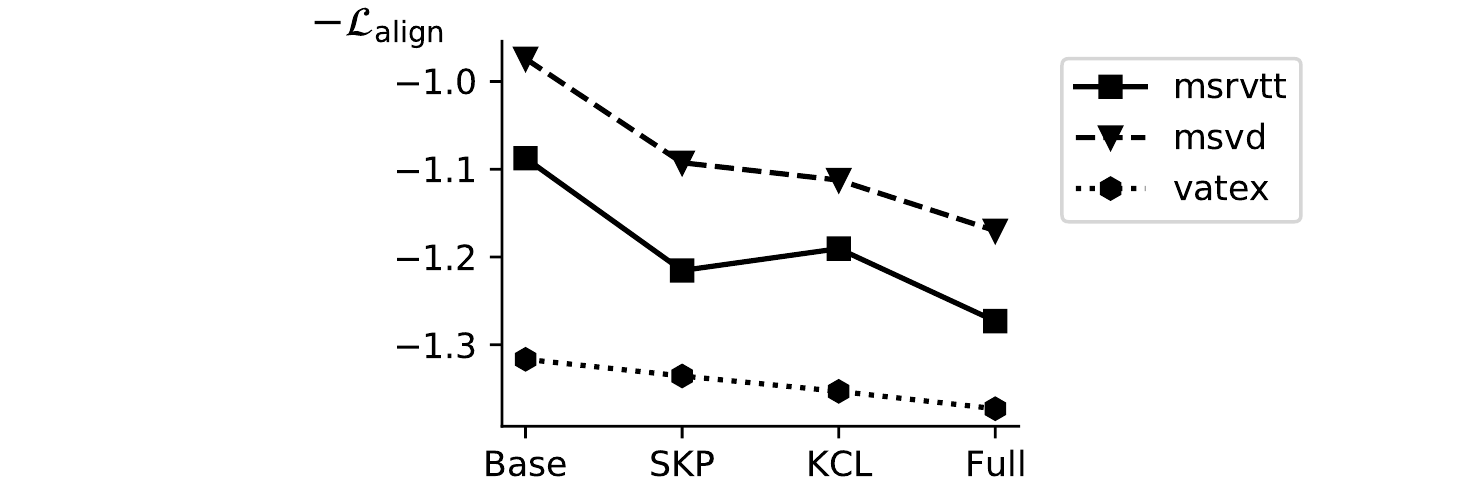} \\
    (a) Alignment between text and videos. \\
    \includegraphics[width=.45\textwidth]{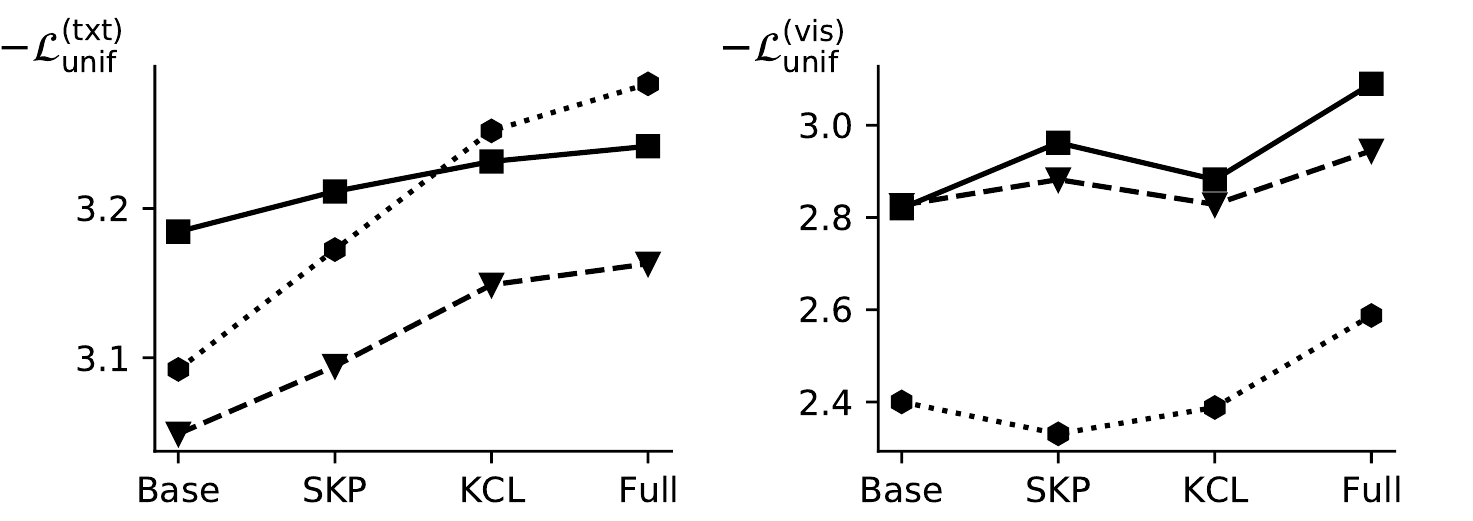} \\
    (b) Uniformity of text (left) and videos (right). \\
  \end{tabular}
  \caption{Alignment and uniformity metrics.}
  \label{fig:align_unif}
\end{figure}

We use these metrics to quantitatively analyze the representation distributions of our model.
The lower $\mathcal{L}_{\mathsf{align}}$ and $\mathcal{L}_{\mathsf{unif}}$ (or equivalently as the y axis in Fig.~\ref{fig:align_unif}, the higher $-\mathcal{L}_{\mathsf{align}}$ and $-\mathcal{L}_{\mathsf{unif}}$) indicate higher quality of representation space.
A recent work~\cite{wang2021understanding} identifies that there is a compromise between these two properties.

As shown in Fig.~\ref{fig:align_unif}, we illustrate the zero-shot alignment and uniformity of our model on three retrieval tasks and have several interesting insights as follows:

{\bf (1) Our method promotes the \textit{uniformity} to benefit the cross-modal retrieval.}
As shown in Fig.~\ref{fig:align_unif}~(b) and Tab.~\ref{tab:ablation}~(1), we observe a clear correlation of the {\it uniformity} enhancements and the retrieval performance boosts.
The model seems to find an effective way to improve the separability for retrieval on the hypersphere: optimizing the {\it uniformity} property to help data points better ``spreading'' on the representation space, even at the cost of hurting {\it alignment} to some extent.

{\bf (2) The SKP and KCL regularizations collaborate to work.}
Our SKP task attracts similar videos to learnable semantic centers.
When combined with KCL which efficiently pushes apart cross-modal hard negatives, they together achieve better {\it uniformity} of the representation space, as shown in Fig.~\ref{fig:align_unif}~(b).
A consistent {\it uniformity} improvement is observed across different retrieval datasets.

Overall, our knowledge regularizations encourage the model to fully utilize the representation space for better separability, thus benefiting the cross-modal retrieval tasks.


\section{Conclusions}
\label{sec:concl}

In this work, we propose CLOP, a video-and-language pre-training method with Knowledge Regularizations.
Our method consists of a simple yet effective Structural Knowledge Prediction (SKP) proxy task and a novel sampling approach for Knowledge-Guided Contrastive Learning (KCL) with hard negatives.
The extensive experiments validate the effectiveness of our method, showing that CLOP outperforms the state-of-the-art methods on five downstream tasks.
An in-depth analysis shows evidence that, our knowledge regularizations promote the {\it uniformity} property to form a superior cross-modal representation space.


\bibliographystyle{ACM-Reference-Format}
\bibliography{references}

\end{document}